\documentclass[letterpaper]{article} % DO NOT CHANGE THIS
\usepackage[submission]{2026}  % DO NOT CHANGE THIS
\usepackage{times}  % DO NOT CHANGE THIS
\usepackage{helvet}  % DO NOT CHANGE THIS
\usepackage{courier}  % DO NOT CHANGE THIS
\usepackage[hyphens]{url}  % DO NOT CHANGE THIS
\usepackage{graphicx} % DO NOT CHANGE THIS
\usepackage{natbib}  % DO NOT CHANGE THIS AND DO NOT ADD ANY OPTIONS TO IT
\usepackage{caption} % DO NOT CHANGE THIS AND DO NOT ADD ANY OPTIONS TO IT
\usepackage{graphicx}
\usepackage{subcaption}
\usepackage{algorithm}
\usepackage{algorithmic}
\usepackage{booktabs} % for \usepackage{amsmath}
\usepackage{amsmath}
\usepackage{amssymb}
\usepackage{mathtools}
\usepackage{amsthm}
\usepackage{multirow}
\usepackage{xcolor}

\newcommand{\undl}[1]{\underline{#1}}
\newcommand{\tts}[1]{\text{#1}^*}
\newcommand{\ttt}[1]{\text{#1}}
\usepackage{newfloat}
\usepackage{listings}
\DeclareCaptionStyle{ruled}{labelfont=normalfont,labelsep=colon,strut=off} % DO NOT CHANGE THIS
\floatstyle{ruled}
\newfloat{listing}{tb}{lst}{}
\floatname{listing}{Listing}
\title{Unified Flow Matching for Long Horizon Event Forecasting}
\author{
    Written by Press Staff\textsuperscript{\rm 1}\thanks{With help from the Publications Committee.}\\
    Style Contributions by Pater Patel Schneider,
    Sunil Issar,\\
    J. Scott Penberthy,
    George Ferguson,
    Hans Guesgen,
    Francisco Cruz\equalcontrib,
    Marc Pujol-Gonzalez\equalcontrib
}
\affiliations{
    \textsuperscript{\rm 1}Association for the Advancement of Artificial Intelligence\\
    1101 Pennsylvania Ave, NW Suite 300\\
    Washington, DC 20004 USA\\
}

\usepackage{bibentry}
\begin{document}

\maketitle

\begin{abstract}
Modeling long horizon marked event sequences is a fundamental challenge in many real-world applications, including healthcare, finance, and user behavior modeling. Existing neural temporal point process models are typically autoregressive, predicting the next event one step at a time, which limits their efficiency and leads to error accumulation in long-range forecasting. In this work, we propose a unified flow matching framework for marked temporal point processes that enables non-autoregressive, joint modeling of inter-event times and event types, via continuous and discrete flow matching. By learning continuous-time flows for both components, our method generates coherent long horizon event trajectories without sequential decoding. We evaluate our model on six real-world benchmarks and demonstrate significant improvements over autoregressive and diffusion-based baselines in both accuracy and generation efficiency.
\end{abstract}

\section{Introduction}
Event sequences are ubiquitous in the real world, where phenomena unfold as discrete events occurring asynchronously over continuous time. Marked Temporal Point Processes (MTPPs) \citep{daley2003introduction} offer a principled mathematical framework for modeling such data, capturing both the timing and type of events. MTPPs have been widely applied across various domains including E-commerce, social networks, finance, and healthcare to model event data and capture domain specific temporal dynamics. Traditionally, parametric models such as the Hawkes process \citep{hawkes1971point} have been used to model event sequences, and they rely on carefully designed intensity functions to capture temporal dynamics and interactions among event types. More recently, neural MTPPs have emerged as powerful alternatives, leveraging deep learning to model complex, non-linear dependencies in irregular event streams. Notable examples include RNN-based models \citep{du2016recurrent,omi2019fully, mei2016neural}, and attention-based approaches such as the Transformer Hawkes Process \citep{zuo2020transformer} and Self-Attentive Hawkes Process \citep{zhang2020self}. These models have demonstrated state-of-the-art performance on next event forecasting task across various applications. %For multi-event or long horizon event forecasting, researchers \cite{xue2022hypro, deshpande2021long} have focused on tuning attention mechanisms.
For multi-event or long horizon forecasting, researchers have proposed diverse approaches beyond attention mechanisms, including hybrid energy-based models \cite{xue2022hypro} and dual event-count frameworks \cite{deshpande2021long} to improve coherence in predictions.

However, all above adopt an autoregressive modeling paradigm. While effective for short horizon forecasting, such approaches often struggle with generating long event sequences due to error accumulation and computational inefficiency. To address this, recent efforts have explored non-autoregressive generative models for event sequences. Diffusion-based methods, such as Add-and-Thin \citep{ludke2023add} and CDiff \citep{zeng2024interacting}, have demonstrated promise in generating long-range event streams. Inspired by advances in computer vision, flow matching \citep{lipmanflow} has emerged as a compelling alternative, offering efficient simulation by learning continuous-time flow dynamics. %Notably, recent preprints~\citep{mukherjee2025adiff4tpp,kerrigan2024eventflow} have proposed flow matching frameworks for (conditional) event generation. However, these approaches are limited to either unmarked event sequences or modeling only latent representations of time and mark, rather than their explicit joint dynamics. The latter case limits the model’s ability to capture their explicit dependencies, reduces interpretability, and restricts control during generation. 
Notably, \textit{concurrent} preprints~\citep{mukherjee2025adiff4tpp,kerrigan2024eventflow} have proposed flow matching frameworks for (conditional) event generation. However, these approaches are limited to either unmarked event sequences or modeling only latent representations of time and mark, rather than capturing their explicit joint dynamics. The latter limitation hinders the model's ability to learn meaningful dependencies between timing and event types, reduces interpretability, and restricts control during generation.

% In this paper, we introduce the \textit{first} unified flow matching framework for long-horizon marked event forecasting, which jointly models both temporal evolution and mark dynamics. Our approach enables efficient simulation of realistic future trajectories without relying on autoregressive decoding. We evaluate our model on six real-world datasets and demonstrate superior performance in generating plausible long-range event sequences.
% In this paper, we introduce, to the best of our knowledge, the \textit{first unified flow matching framework} for long-horizon marked event forecasting, jointly explicitly modeling both temporal evolution and event type dynamics, along with a novel sampling algorithm for marked event generation. Our approach enables efficient simulation of realistic future trajectories without the need for autoregressive decoding. Extensive evaluations on six real-world datasets demonstrate the framework’s superior ability to generate plausible long-range event sequences.
To this end, we introduce, to the best of our knowledge, the \textit{first unified flow matching framework} for long-horizon marked event forecasting. Our method explicitly models the joint dynamics of inter-event times and event types through coupled flows, and introduces a novel sampling algorithm tailored for efficient marked event generation. This unified approach enables non-autoregressive simulation of realistic long-range event sequences with high temporal and categorical fidelity. Extensive evaluations on six real-world datasets demonstrate the framework’s strong performance in generating plausible long-range event sequences.

\section{Related Work}
\paragraph{Neural Marked Temporal Point Process.} 
The use of neural networks has significantly expanded the modeling capacity of MTPPs. Unlike classical parametric models, neural MTPPs can capture complex and nonlinear dependencies among events, enabling more flexible representations of event dynamics \cite{shchur2021neural}. Early approaches, such as recurrent neural network-based models, encode historical dependencies through hidden state embeddings \cite{du2016recurrent,mei2016neural,omi2019fully,shchur2019intensity}, while more recent methods employ attention mechanisms and transformers to model interactions between events over time \cite{yang2021transformer,zhang2020self,zuo2020transformer,gu2021attentive, shou2023influence,shou2023concurrent}. While some models learn the conditional intensity function directly, others bypass it altogether by modeling inter-event time distributions or cumulative intensities. These neural approaches offer improved expressiveness and have demonstrated strong performance across a range of event modeling tasks.

\paragraph{Long Horizon Event Forecasting.} 

Several MTPP models explicitly target long horizon event forecasting, but they predominantly rely on sequential generation mechanisms and autoregressive modeling, which learn the conditional distribution of the next event given the history. HYPRO~\citep{xue2022hypro} generates multiple candidate event sequences and introduces a selection module to identify the most plausible trajectory. %Dual-TPP~\citep{deshpande2021long} proposes a hierarchical architecture combined with a ranking objective to better estimate the number of future events within fixed intervals. 
Dual-TPP~\citep{deshpande2021long} introduces a two-component architecture, combining an event model for fine-grained predictions and a count model for aggregated intervals, with a joint optimization objective to improve long horizon forecasting accuracy. More recently, non-autoregressive approaches have gained attention for long horizon generation. In the unmarked setting, Lüdke et al.~\citep{ludke2023add} introduce the Add-and-Thin diffusion model, which treats inter-event time sequences as samples from a Poisson-based conditional distribution. However, this method does not account for event types and is limited to modeling time alone. To overcome this limitation, CDiff~\citep{zeng2024interacting} proposes a diffusion-based framework that jointly models the generative process of timestamps and event types. It employs two coupled denoising diffusion processes: a categorical diffusion for event types and a real-valued diffusion for inter-event times, whose interaction allows the model to capture dependencies between timing and event types, and thus resulting in more coherent and realistic event sequence generation.

%In contrast, our work targets the *joint modeling* of time intervals and event types in multivariate settings. We directly learn a continuous-time generative process for marked event sequences, enabling efficient and accurate long-horizon simulation. For completeness, we compare our method to a modified Add-and-Thin model augmented with a naive mark predictor in the Appendix.

\paragraph{Flow Matching.}
Flow Matching (FM) for continuous variables~\citep{lipmanflow} is a recently proposed framework for generative modeling that learns to transport a simple base distribution (e.g., Gaussian noise) to a complex data distribution via a deterministic flow governed by an ordinary differential equation (ODE). Specifically, FM trains a learnable vector field $\mathbf{v}_\theta(x, t)$ to approximate a target vector field $\mathbf{v}_t(x)$ that governs the evolution of intermediate distributions $\{p_t(x)\}_{t \in [0,1]}$ along a probability path. The sample dynamics are defined by the ODE $\frac{d}{dt} \phi_t(x) = \mathbf{v}_\theta(\phi_t(x), t)$, where $\phi_t(x)$ denotes the trajectory of a sample over time and $\phi_0(x) = x $. The training objective minimizes the expected squared deviation between the learned and target vector fields across time and samples. Compared to diffusion models, FM avoids iterative denoising and stochastic sampling, offering a more efficient alternative for learning continuous-time generative processes. Recent work has demonstrated its effectiveness in domains such as image generation~\citep{lipmanflow}. FM has been extended to model discrete (categorical) variables by leveraging continuous-time Markov chains and Kolmogorov forward equations to model dynamics in discrete spaces. This has enabled applications in domains such as vocabulary and word token modeling in natural language processing \cite{gat2024discrete}, as well as protein co-design~\citep{campbell2024generative}.

%and long-horizon event sequence modeling~\citep{mukherjee2025adiff4tpp,kerrigan2024eventflow}. In this paper, we extend flow matching to the setting of marked temporal point processes, proposing a unified framework that jointly models both inter-event times and event types.

\section{Problem Definition}

We consider a marked event sequence consisting of event timestamps and associated labels, represented as \( S = \{(t_i, y_i)\}_{i=1}^n \), where \( t_i \in \mathbb{R}_+ \) denotes the time of the \( i^{\text{th}} \) event and \( y_i \in \mathbb{L} \) is its corresponding event type, drawn from a finite label set \( \mathbb{L} \) with cardinality $M$. To ensure translation invariance and facilitate modeling, we convert absolute timestamps into inter-event times by defining \( x_i = t_i - t_{i-1} \) for \( i = 1, \dots, n \), with the convention \( t_0 = 0 \). Each sequence is then represented as \( S = \{(x_i, y_i)\}_{i=1}^n \).

In the long horizon event forecasting setting, the goal is to generate a future subsequence \( S_f = \{(x_i, y_i)\}_{i=c+1}^n \), conditioned on a prefix (history context) \( S_c = \{(x_i, y_i)\}_{i=1}^c \). That is, given the first \( c \) events, the model aims to forecast a fixed number of future events, the next \( n - c \) events (i.e. 20). Typical autoregressive MTPPs model the joint $p(S_f | S_c) = \Pi_{i=c+1}^n p^*(x_i,y_i|S_c)$ where $*$ indicates all events after the context and prior to event $i$. Such autoregressive nature causes error accumation due to the reliance of predicted time and mark at each step. 

\section{Methodology}
We propose a novel end-to-end \undl{U}nified \undl{F}low \undl{M}atching framework for modeling marked \undl{T}emporal \undl{P}oint \undl{P}rocesses (UFM-TPP), tailored to long horizon event forecasting. Given a context sequence of past interevent times and marks, our model learns to forecast a sequence of future event timings and types by learning to transporting samples from simple noise distributions to a (potentially complex) target sample distribution, using flow-matching principles. 

\subsection{Unified Flow Matching}
\begin{figure}[h!]
    \centering
    \includegraphics[width=0.8\linewidth]{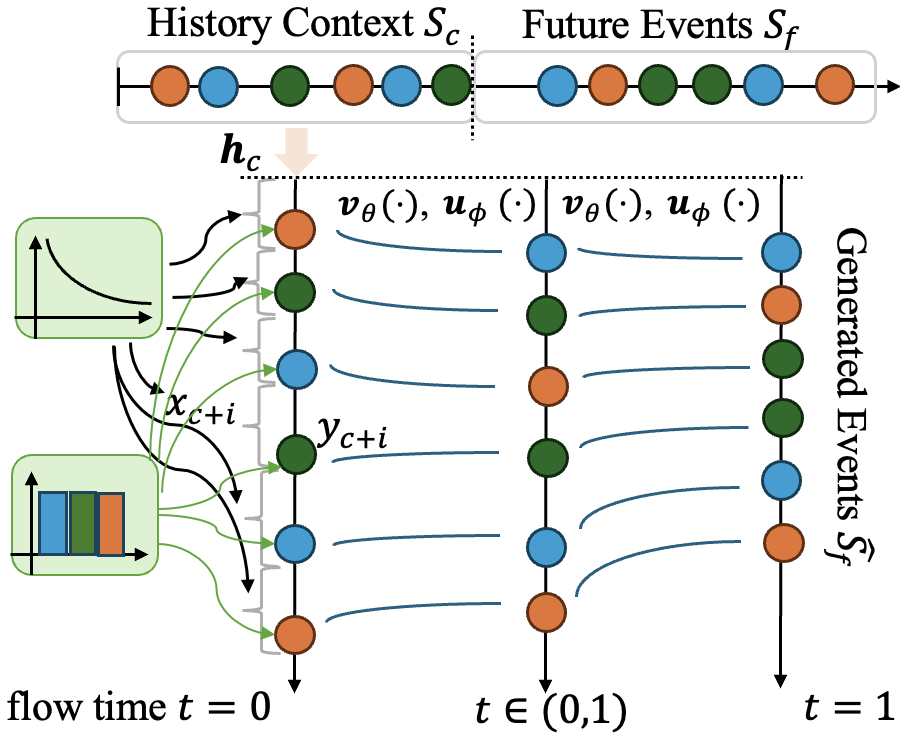}
    \caption{
    Overview of the proposed model UFM-TPP. Given a history context \( S_c \), we aim to generate a future sequence \( S_f \) of event tuples \((x_{c+i}, y_{c+i})\) using a coupled flow model.
    At flow time \( t = 0 \), we initialize from noise drawn from simple base distributions: exponential for inter-event times and uniform distribution over discrete categories for marks. A shared context encoder produces a representation \( \mathbf{h}_c \), which conditions both the continuous vector field \( \mathbf{v}_\theta \) and the discrete logits \( \mathbf{u}_\phi \). These components jointly guide the evolution of time and mark trajectories toward realistic events at \( t = 1 \). Blue lines indicate continuous-time flow interpolation; colored circles represent discrete event types.
    % A shared context encoder produces a representation \( \mathbf{h}_c \), conditioning both the continuous vector field \( \mathbf{v}_\theta \) and discrete logits \( \mathbf{u}_\phi \). 
    % These jointly drive the evolution of time and mark trajectories toward realistic events at \( t = 1 \). 
    % Blue lines denote the continuous-time flow interpolation, while colored circles represent discrete event types.
    }
    \label{fig:ufm_overview}
\end{figure}

Our method UFM-TPP jointly models the evolution of inter-event times ${x_{i}}'s$ and event types ${y_{i}}'s$ for $i = \{c+1, c+2,...,n\}$ using two coupled flow dynamics: a continuous flow for timestamps and a discrete flow for categorical marks. This enables efficient, non-autoregressive generation of multi-typed event sequences. To avoid error accumulation, we make a simplifying assumption in our setting:
\begin{equation}
\label{eqn: assm}
  \Pi_{i=c+1}^n p^*(x_i,y_i|S_c) \approx \Pi_{i=c+1}^n p(x_i,y_i|S_c)  
\end{equation} where  we drop the dependence of prior events and thus all event tuples become IID given history context. An visual illustration of our model is shown in Fig.\ref{fig:ufm_overview}.

\paragraph{Context Encoding.}
To condition future event generation on past observations, we encode the event history using a sequence encoder. Given a context window consisting of inter-event times $x_i$'s and corresponding event types $y_i$'s, we first embed each modality: inter-event times are passed through a time embedding network, and event types are embedded via a trainable categorical embedding. The resulting vectors are concatenated to form a unified event representation:
$
\mathbf{z}_i = [\texttt{MarkEmbed}(y_i) \,\|\, \texttt{TimeEmbed}( x_i)], \quad \text{for } i = \{1, \dots, c\}.
$
We then use a sequence model  such as a recurrent neural network (RNN) with parameter $\psi$ to encode the sequence \( \{\mathbf{z}_1, \dots, \mathbf{z}_c\} \) into a fixed-length context vector $\mathbf{h}_c \in \mathbb{R}^d$ for some dimension $d$. %To handle variable-length input sequences, we apply sequence packing based on true context lengths during training. 
This context vector serves as a summary of the historical prefix $S_c$ and is used to condition the flow model at each timestep. The entire encoding pipeline is differentiable, allowing our model to be trained end-to-end jointly with the flow-based generator.

\paragraph{Continuous Flow for Inter-Event Times.} Inspired by flow matching~\citep{lipmanflow}, we propose a novel continuous-time framework for modeling inter-event times in marked temporal point processes. Given an inter-event time \( x_i \) from future target $S_f$, we define a linear interpolation between a noise sample \( x_i^0 \sim p_0(x) = \text{Exp}(\lambda) \) and the ground truth value \( x_i^1 = x_i \) as:
\begin{equation}
   x_i(t) = (1 - t) x_i^0 + t x_i^1, \quad \text{for} \quad t \in [0, 1].
\end{equation}
We choose the exponential distribution as the base due to its connection to the homogeneous Poisson process, where \( \lambda \) denotes the event rate or the reciprocal of interevent times. This rate can be set manually or estimated from the context window \( S_c \). This ensures that as \( t \to 1 \), the distribution of \( x_i(t) \) smoothly contracts to the ground truth value \( x_i^1 \), enabling a well-behaved flow from noise to data. 

%Since \( x_i(t) \) is an affine transformation of an exponential variable, it follows a shifted and scaled exponential distribution with closed-form density:
% \[
% f_{x_i(t)}(x) = 
% \begin{cases}
% \lambda \exp\left( -\lambda \cdot \frac{x - t x_i^1}{1 - t} \right) \cdot \frac{1}{1 - t}, & x \geq t x_i^1 \\
% 0, & \text{otherwise}.
% \end{cases}
% \]

\paragraph{Discrete Flow for Event Types.} Inspired by \cite{gat2024discrete}, we propose a discrete flow construction for event types (marks). For each event type  \( y_i \). Let \( y_i^1 = y_i \) denote the ground-truth mark, and let \( y_i^0 \sim p_0(y) = \text{Categorical}(\pi = \frac{1}{C}) \) be a noise sample from a uniform categorical distribution or can be similarly estimated from summary statics of $S_c$. We define a stochastic interpolation between \( y_i^0 \) and \( y_i^1 \). %using a time-dependent Bernoulli switch:
% \[
% y_i(t) =
% \begin{cases}
% y_i^1, & \text{with probability } t \\
% y_i^0, & \text{with probability } 1 - t.
% \end{cases}
% \]
This construction induces a time-dependent mixture distribution over marks:
\begin{equation}
y_i(t) = (1 - t) \cdot p_0(y_i) + t \cdot \delta_{y_i^1}(y_i),
\end{equation}
where \( \delta_{y_i^1}(\cdot) \) is the Dirac delta centered at the true mark \( y_i^1 \). As time \( t \) increases from 0 to 1, this interpolation gradually shifts probability mass from the base distribution \( p_0(y) \) to the target label, providing a smooth discrete transition path suitable for training flow-based models over categorical variables.

\paragraph{Joint Training.}
Instead of modeling inter-event times and event types separately, we propose to jointly capture their dependencies by modeling the conditional distributions \( p(x_i \mid y_i, \mathbf{h}_c) \) and \( p(y_i \mid x_i, \mathbf{h}_c) \), conditioned on the context representation vector. This joint formulation enables the model to learn the dynamics of time-mark event tuples in a principled and flexible manner. To model inter-event times, we train a neural vector field \( \mathbf{v}_\theta(x_i(t), y_i(t), t, \mathbf{h}_c) \) to predict the time derivative \( \frac{d}{dt} x_i(t) \) along the flow trajectory. Following the flow matching framework with an optimal transport interpretation~\citep{tong2024improving}, we minimize the squared error between the predicted and target derivatives:
{\small
\begin{equation}
\label{eqn:time}
\mathcal{L}_{\text{time}} = \mathbb{E}_{t \sim \mathcal{U}[0,1], x_i^0 \sim p_0(x), x_i^1 \sim S_f} \left[ \sum_{i=c+1}^{n} \left\| \mathbf{v}_\theta(\cdot) - (x_i^1 - x_i^0) \right\|^2 \right]
\end{equation}
}
where \( \mathbf{v}_\theta(\cdot) \) is shorthand for \( \mathbf{v}_\theta(x_i(t), y_i(t), t, \mathbf{h}_c) \). We assume the independent time coupling between source and target $(x_i^0, x_i^1) \sim \Pi(X^0 ,X^1)$ for efficiency as indicated by $x_i^0 \sim p_0(x), x_i^1 \sim S_f$ in Eqn. \ref{eqn:time}.  To model event types, we estimate the conditional distribution \( p(y_i \mid x_i, \mathbf{h}_c) \) to capture interactions between temporal and categorical components. We adopt a discrete flow matching approach, which constructs a probability path using a rate matrix that governs transition dynamics in label space~\citep{gat2024discrete,campbell2024generative}. While in principle one could directly model the time derivative \( \frac{d}{dt} p(y_i) \), in practice we instead model this as a denoising process using a cross-entropy objective. The model learns to predict the clean label \( y_i^1 \) from noisy and intermediate input \( y_i(t) \). Our neural model outputs logits \( \mathbf{u}_\phi(x_i(t), y_i(t), t, \mathbf{h}_c) \in \mathbb{R}^ M \), and we minimize:
{\scriptsize
\begin{equation}
\label{eqn:lab}
\mathcal{L}_{\text{mark}} = \mathbb{E}_{t \sim \mathcal{U}[0,1],y_i^0 \sim p_0(y_i), y_i^1 \sim S_f} \left[ \sum_{i=c+1}^{n} \text{CE}(\text{softmax}(\mathbf{u}_\phi(\cdot)), y_i^1) \right]
\end{equation}
}
where \( \mathbf{u}_\phi(\cdot) \) denotes the same functional input as above (i.e., $ \mathbf{u}_\phi(x_i(t), y_i(t), t, \mathbf{h}_c)$). Similarly, we assume the independent mark coupling between source and target $(y_i^0, y_i^1) \sim \Pi(Y^0 ,Y^1)$ as  $y_i^0 \sim p_0(y), y_i^1 \sim S_f$ for efficiency shown in Eqn. \ref{eqn:lab}. By jointly training the continuous and discrete flows, both conditioned on shared context, our model learns to simulate long horizon marked event sequences in a non-autoregressive and coherent manner. The final training objective is:
\begin{equation}
\mathcal{L}_{\text{total}} = \mathcal{L}_{\text{time}} + \alpha \cdot \mathcal{L}_{\text{mark}},
\end{equation}
where \( \alpha \) balances the two objectives. Empirically we find \( \alpha = 1 \) works well across datasets. The full model $\mathcal{M}$ consists of a continuous vector field \( \mathbf{v}_\theta \), a discrete classifier \( \mathbf{u}_\phi \), and a context encoder \( \text{RNN}_\psi \).

\begin{algorithm}[htbp]
\caption{Joint Flow Matching Sampling for Batchwise Long Horizon Event Generation}
\label{alg:joint-sampling}
\begin{algorithmic}[1]
\REQUIRE Target length $L$, steps $S$, base rate $\lambda$, base mark distribution $\pi_0$, context $S_c$
\ENSURE Future inter-event times $x \in \mathbb{R}^{B \times L}$ and marks $y \in \{1,\dots,\mathbb{L}\}^{B \times L}$ 
\STATE Initialize step size $h \gets 1/S$, time $t \gets \mathbf{0}_B$
\STATE Sample initial noise: $x \sim \text{Exp}(\lambda)$, $y \sim \text{Categorical}(\pi_0)$
\FOR{$s = 1$ to $S$}
    \STATE Compute drift and logits: $(v_0, \text{logits}_0) \gets \text{Model}(x, y, t, S_c)$
    \STATE Midpoint state: $x_{\text{mid}} \gets \text{max}(x + \frac{h}{2} \cdot v_0, \epsilon)$
    \STATE Midpoint time: $t_{\text{mid}} \gets t + \frac{h}{2}$
    \STATE $(v_{\text{mid}}, \text{logits}_{\text{mid}}) \gets \text{Model}(x_{\text{mid}}, y, t_{\text{mid}}, S_c)$
    \STATE Update inter-event time: $x \gets \text{max}(x + h \cdot v_{\text{mid}},\epsilon)$
    \STATE Compute probabilities: $p \gets \text{softmax}(\text{logits}_0)$
    % \STATE Compute one-hot: $\text{one\_hot} \gets \text{OneHot}(y)$
    \STATE Velocity in mark space: $u \gets (p - \mathbf{1}_{y}) / (1 - t)$
    \STATE Update mark distribution: $p_{\text{new}} \gets \text{max}(\mathbf{1}_{y_t} + h \cdot u, \epsilon)$
    \STATE Sample new mark: $y \sim \text{Categorical}(p_{\text{new}})$
    \STATE Update time: $t \gets t + h$
\ENDFOR
\RETURN $(x, y)$
\end{algorithmic}
\end{algorithm}

\paragraph{Sampling.}
We propose a novel joint sampling algorithm for long horizon marked event generation that evolves both inter-event times and event types from initial noise toward realistic future events using learned flow model $\mathcal{M}$ via \textit{constrained} generation. The sampling begins by drawing noise samples from simple base distributions: \( x_i^0 \sim \text{Exp}(\lambda) \) for inter-event times and \( y_i^0 \sim \text{Categorical}(\pi_0) \) for marks. These serve as the starting point for simulating a deterministic trajectory governed by learned vector fields. Since each event tuple $(x_i,y_i|S_c)$ is IID we drop the subscript $i$ for ease of notation to describe the sampling process.    

At each timestep \( t \in [0,1] \), the algorithm updates the state using a second-order midpoint method \cite{lipman2024flow} for continuous time and a discrete flow update for event types . The continuous trajectory is updated as:
\begin{equation}
\begin{aligned}
x_{\text{mid}} &= \text{max}(x_t + \tfrac{h}{2} \cdot \mathbf{v}_{\theta^*}(x_t, y_t, t, \mathbf{h}_c),\epsilon), \\
x_{t+h} &= \text{max}(x_t + h \cdot \mathbf{v}_{\theta^*}(x_{\text{mid}}, y_t, t + \tfrac{h}{2}, \mathbf{h}_c),\epsilon).
\end{aligned}
\end{equation}

where \( \mathbf{v}_{\theta^*} \) is the learned vector field for time evolution and $h$ is the step size. It is worth noting that we place an efficient projection to the positive orthant ensure \textit{positivity} of the interevent times $x$ at each generation step.  

For the discrete mark, we compute the softmax probabilities \( p_t = \text{softmax}(\mathbf{u}_{\phi^*}(x_t, y_t, t, \mathbf{h}_c)) \), where \( \mathbf{u}_{\phi^*} \) denotes the predicted logits from the trained model. Letting \( \mathbf{1}_{y_t} \) be the one-hot encoding of the current mark, we compute the velocity in the probability simplex and updated distribution:
\begin{equation}
u = \frac{p_t - \mathbf{1}_{y_t}}{1 - t}, \quad 
p_{t+h} = \text{max}(\mathbf{1}_{y_t} + h \cdot u, \epsilon)
\end{equation}
where we impose clamp with threshold \( \epsilon>0 \) ensures numerical stability. Lastly we sample $y_{t+h} \sim \text{Categorical}(p_{t+h})$ parameterized by the updated probability simplex. 

By iteratively applying these updates over a fixed number of steps, the algorithm transforms samples from the base distribution into coherent, long horizon event sequences. The resulting trajectories jointly model both the temporal and categorical components in a non-autoregressive manner, conditioned on the encoded context \( S_c \). The full algorithm is described in Alg. \ref{alg:joint-sampling}.

\begin{table*}[h!]
    \caption{$\textbf{OTD}$, $\textbf{RMSE}_{y}$, $\textbf{RMSE}_{x}$ and \textbf{sMAPE} of real-world datasets reported in mean $\pm$ s.d. Best are in bold, the next best is underlined. *indicates statistical significance w.r.t. the best method using Welch's t-test (p $<$ 0.05).}
    \vskip 0.1in
    \centering\resizebox{\linewidth}{!}{\begin{tabular}{lcccc|cccc} \toprule
    &\multicolumn{4}{c}{\textbf{Taxi}}&\multicolumn{4}{c}{\textbf{Taobao}}\\ 
        & $\textbf{OTD}$ & $\textbf{RMSE}_{y}$ &$\textbf{RMSE}_{x}$ & \textbf{sMAPE} & $\textbf{OTD}$ & $\textbf{RMSE}_{y}$&$\textbf{RMSE}_{x}$ & \textbf{sMAPE}\\ \midrule
    \textbf{HYPRO}      & $\text{21.653 $\pm$ 0.163}^*$ & $\text{\underline{1.231 $\pm$ 0.015}}^*$ & $\text{0.372 $\pm$ 0.004}^*$ & $\text{93.803 $\pm$ 0.454}^*$ & $\tts{44.336 $\pm$ 0.127}$  &$\text{2.710 $\pm$ 0.021}^*$ & $\text{{0.594 $\pm$ 0.030}}^*$ & $\text{{134.922 $\pm$ 0.473}}^*$ \\
     \textbf{Dual-TPP}      & ${24.483 \pm 0.383}^*$ & $\text{1.353 $\pm$ 0.037}^*$ & $\text{0.402 $\pm$ 0.006}^*$ & $\text{95.211 $\pm$ 0.187}^*$ & $\text{47.324 $\pm$ 0.541}^*$ & $\text{3.237 $\pm$ 0.049}^*$ & $\text{0.871 $\pm$ 0.005}^*$ & $\text{141.687 $\pm$ 0.431}^*$\\ 
      \textbf{Attnhp}     &  ${24.762 \pm 0.217}^*$  &$\text{1.276  $\pm $ 0.015}^*$ & ${0.430 \pm 0.003}^*$& $\text{97.388 $\pm$  0.381}^*$ &  $\text{45.555 $\pm$ 0.345}^*$  & $\tts{2.737 $\pm$ 0.021}$ & $\text{0.708 $\pm$ 0.010}^*$ & $\text{134.582 $\pm$ 0.920}^*$ \\ 
       \textbf{NHP}      & $\text{25.114 $\pm$ 0.268}^*$ &$ {1.297 \pm 0.019}^*$ & ${0.399 \pm 0.040}^*$ & $\text{96.459 $\pm$ 0.521}^*$ &    $\text{48.131 $\pm$ 0.297}^*$   &$\text{3.355  $\pm$ 0.030}^*$ & $\text{0.837 $\pm$ 0.009}^*$ & $\text{137.644 $\pm$ 0.764}^*$\\  
      \textbf{LNM}     &  $\tts{24.053 $\pm$ 0.609}$  &$\tts{1.364 $\pm$ 0.032}$ & $\tts{0.384 $\pm$ 0.005}$& $\tts{95.719 $\pm$ 0.779}$ &  $\tts{45.757 $\pm$ 0.287}$  & $\tts{3.193 $\pm$ 0.043}$ & $\tts{0.575 $\pm$ 0.012}$ & $\tts{{ 127.436 $\pm$ 0.606}}$ \\ 
       \textbf{TCDDM}      & $\tts{22.148 $\pm$ 0.529}$ & $\tts{  1.309 $\pm$ 0.030}$ & $\tts{  0.382 $\pm$ 0.019}$  & $\tts{ 90.596 $\pm$ 0.574}$  &    $\tts{45.563 $\pm$ 0.889}$    & $\tts{ 2.850 $\pm$ 0.058}$  & $\tts{ 0.569 $\pm$ 0.015}$  & $\tts{ \undl{126.512 $\pm$ 0.491}}$ \\      
         \textbf{CDiff}    & \textbf{21.013  $\pm$  0.158} & \textbf{1.131 $\pm$ 0.017} & $\tts{\undl{0.351 $ \pm$ 0.004}}$ & $\tts{\undl{87.993 $\pm$ 0.178}}$ &   $\tts{44.621 $\pm$ 0.139}$ &$\tts{\undl{{2.653 $\pm$ 0.022}}}$ & $\tts{ \undl{{0.551 $\pm$ 0.002}}}$ & \textbf{125.685 $\pm$ 0.151} \\
        \midrule
         \textbf{UFM-TPP}    & \text{\undl{21.282 $\pm$  1.124}}   & $\tts{{1.254 $\pm$ 0.067}}$ & \textbf{{0.298 $ \pm$ 0.008}}& \textbf{81.544 $\pm$ 1.550} &   \textbf{40.552 $\pm$ 0.648}  &\textbf{{2.169 $\pm$ 0.030}} & \textbf{{0.436 $\pm$ 0.014}} & $\tts{161.212 $\pm$ 1.387}$ \\\bottomrule
         &\multicolumn{4}{c}{\textbf{StackOverflow}}&\multicolumn{4}{c}{\textbf{Retweet}} \\
         & $\textbf{OTD}$ & $\textbf{RMSE}_{y}$&$\textbf{RMSE}_{x}$ & \textbf{sMAPE} & $\textbf{OTD}$ & $\textbf{RMSE}_{y}$&  $\textbf{RMSE}_{x}$ & \textbf{sMAPE}\\ \midrule
         \textbf{HYPRO} & $\tts{{42.359 $\pm$ 0.170}}$ & \undl{1.140 $\pm$ 0.014} & $\text{1.554 $\pm$ 0.010}^*$ &  $\text{ 110.988 $\pm$  0.559}^*$ & $\text{61.031 $\pm$ 0.092}^*$ &$\text{2.623 $\pm$ 0.036}^*$ & $\text{30.100 $\pm$ 0.413}^*$  & $\tts{106.110 $\pm$ 1.505}$\\
         \textbf{Dual-TPP} & $\tts{41.752 $\pm$ 0.200}$ &\textbf{1.134 $\pm$ 0.019} & $\text{1.514 $\pm$ 0.017}^*$ &   $\text{ 117.582  $\pm$ 0.420}^*$ &    $\text{ 61.095 $\pm$ 0.101}^*$  &$\text{2.679 $\pm$ 0.026}^*$ & $\tts{28.914 $\pm$ 0.300}$  & $\tts{106.900 $\pm$ 1.293}$\\
         \textbf{AttNHP} & $\text{42.591 $\pm$ 0.408}^*$  &1.142 $\pm$ 0.011 & $\tts{{1.340 $\pm$ 0.006}}$ & $\tts{108.542 $\pm$  0.531} $      &    $\tts{\undl{ 60.634 $\pm$ 0.097}}$         & $\tts{2.561 $\pm$ 0.054}$ & $\text{28.812 $\pm$ 0.272}^*$ & $\text{107.234 $\pm$ 1.293}^*$ \\
         \textbf{NHP} & $\text{43.791  $\pm$ 0.147}^*$ &$\text{1.244  $\pm$ 0.030}^*$ & $\text{1.487 $\pm$ 0.004}^*$ &      $\text{ 116.952 $\pm$ 0.404}^*$       &      $\tts{60.953 $\pm$ 0.079} $          &$\text{2.651  $\pm$ 0.045}^*$ & $\tts{27.130 $\pm$ 0.224} $&$\text{ 107.075 $\pm$ 1.398}^*$\\ 
          \textbf{LNM} & $\tts{46.280 $\pm$ 0.892}$   & $\tts{  1.447 $\pm$ 0.057}$  & $\tts{ 1.669 $\pm$ 0.005}$  & $\tts{  115.122 $\pm$ 0.627}$        &    $\tts{61.715 $\pm$ 0.152}$         & $\tts{2.776 $\pm$ 0.043}$  & $\tts{27.582 $\pm$ 0.191}$  & $\tts{106.711 $\pm$ 1.615}$  \\
         \textbf{TCDDM} & $\tts{42.128 $\pm$ 0.591}$  & $\tts{ 1.467 $\pm$ 0.014}$  & $\tts{ 1.315 $\pm$ 0.004}$  & $\tts{107.659 ± 0.934}$     &   \textbf{60.501 $\pm$ 0.087}      & $\tts{\undl{2.387 $\pm$ 0.050}}$  & $\tts{27.303 $\pm$ 0.152}$  &$\tts{\undl{{106.048 $\pm$ 0.610}}}$\\ 
         \textbf{CDiff} & $\tts{\undl{41.245 $\pm$ 1.400}}$ & {{1.141 $\pm$ 0.007}} & $\tts{\undl{{1.199 $\pm$ 0.006}}}$&      $\tts{\undl{106.175$\pm$ 0.340}}$   &  $\tts{60.661 $\pm$ 0.101}$   & \textbf{{2.293 $\pm$ 0.034}} & $\tts{\undl{{27.101 $\pm$ 0.113}}} $ & $\tts{106.184 $\pm$ 1.121}$\\
        \midrule
         \textbf{UFM-TPP} & \textbf{39.439 $\pm$ 0.618 } & $\tts{{1.439 $\pm$ 0.173}}$ & \textbf{{1.041 $\pm$ 0.027}}&      \textbf{88.890 $\pm$ 1.086}   &  $\tts{61.163 $\pm$ 0.296}$   & $\tts{3.001 $\pm$ 0.272}$ & \textbf{{22.647 $\pm$ 0.209} }  & \textbf{90.291 $\pm$ 0.579}\\
         \bottomrule
          &\multicolumn{4}{c}{\textbf{MOOC}}&\multicolumn{4}{c}{\textbf{Amazon}} \\
         & $\textbf{OTD}$ & $\textbf{RMSE}_{y}$&$\textbf{RMSE}_{x}$ & \textbf{sMAPE} & $\textbf{OTD}$ & $\textbf{RMSE}_{y}$&  $\textbf{RMSE}_{x}$ & \textbf{sMAPE}\\ \midrule
         \textbf{HYPRO} & $\tts{48.621 $\pm$ 0.352}$ & $\tts{\text{1.169 $\pm$ 0.094}}$ & $\tts{ \undl{0.410 $\pm$ 0.005}}$ & $\textbf{143.045 $\pm$ 7.992}$ & $\tts{\text{38.613 $\pm$ 0.536}}$ & $\tts {\undl{2.007 $\pm$ 0.054}}$ & $\tts{0.477 $\pm$ 0.010}$ & $\tts{82.506 $\pm$ 0.840}$  \\
         \textbf{Dual-TPP} & $\tts{50.184 $\pm$ 1.127}$ & $\tts{1.312 $\pm$ 0.019}$ & $\tts{0.435 $\pm$ 0.006}$ & $\ttt{147.003 $\pm$ 2.908}$ & $\tts{42.646 $\pm$ 0.752}$ & $\tts{2.562 $\pm$ 0.202}$ & $\tts{0.482 $\pm$ 0.012}$ & $\tts{86.453 $\pm$ 2.044}$  \\
         \textbf{AttNHP} & $\tts{49.121 $\pm$ 0.720}$ & $\tts{1.297 $\pm$ 0.049}$ & $\tts{0.420 $\pm$ 0.009}$ & $\ttt{147.756 $\pm$ 4.812}$  & $\ttt{39.480 $\pm$ 0.326}$ & $\tts{2.166 $\pm$ 0.026}$ & $\tts{0.476 $\pm$ 0.033}$ & $\tts{84.323 $\pm$ 1.815}$ \\
         \textbf{NHP} &$\tts{51.277 $\pm$ 1.768}$ & $\tts{1.458 $\pm$ 0.063}$ & $\tts{0.442 $\pm$ 0.007}$ & $\ttt{148.913 $\pm$ 11.628}$  & $\tts{42.571 $\pm$ 0.293}$ & $\tts{2.561 $\pm$ 0.060}$ & $\tts{0.519 $\pm$ 0.023}$ & $\tts{92.053 $\pm$ 1.553}$ \\ 
         \textbf{LNM} & $\tts{52.890 $\pm$ 1.151}$ & $\tts{1.428 $\pm$ 0.061}$ & $\tts{0.454 $\pm$ 0.008}$ & $\ttt{149.987 $\pm$ 16.581}$   & $\tts{43.820 $\pm$ 0.232}$ & $\tts{3.050 $\pm$ 0.286}$ & $\tts{0.481 $\pm$ 0.145}$ & $\tts{90.910 $\pm$ 1.611}$  \\
         \textbf{TCDDM} & $\tts{50.739 $\pm$ 0.765}$ & $\tts{1.407 $\pm$ 0.112}$ & $\tts{0.429 $\pm$ 0.015}$ & \undl{145.745 $\pm$ 11.835} & $\tts{42.245 $\pm$ 0.174}$ & $\tts{2.998 $\pm$ 0.115}$ & $\tts{\text{0.476 $\pm$ 0.111}}$ & $\tts{83.826 $\pm$ 1.508}$  \\ 
         \textbf{CDiff} & $\tts{\undl{47.214 $\pm$ 0.628}}$   & $\textbf{{1.095 $\pm$ 0.048}}$ & $\tts{{0.411 $\pm$ 0.009}}$ & $\ttt{146.361 $\pm$ 14.837}$ & $\tts{\undl{37.728  $\pm$  0.199}}$   & $\tts{{2.091 $\pm$ 0.163}}$ & $\tts{\undl{{0.464 $\pm$ 0.086}}}$ & \undl{81.987 $\pm$ 1.905} \\
        \midrule
         \textbf{UFM-TPP} & $\textbf{46.573 $\pm$ 0.540}$   & $\ttt{\undl{1.130 $\pm$ 0.025}}$ & $\textbf{{0.358 $\pm$ 0.014}}$ & $\tts{174.566 $\pm$ 5.741}$ & \textbf{36.031  $\pm$  0.376}   & \textbf{{1.931 $\pm$ 0.074}} & \textbf{{0.349 $\pm$ 0.003}} & \textbf{80.939 $\pm$ 1.736}\\
         \bottomrule
    \end{tabular}}
    \label{tab:condensed_table}
\end{table*}

\section{Experiments}
% We evaluate our proposed model \textbf{UFM-TPP} on multiple real-world benchmarks for long horizon marked event forecasting. In this set of experiments, we fix the number of future events to predict to \( n - c = 20 \) and extend to 10 and 5 in ablation study. Since event sequences vary in total length, the context length \( c \) also varies accordingly. This setup aligns with the objective of long horizon multi-event forecasting, where models must generate entire future sequences rather than just the next event. We impment our model using standard Pytorch code and adapted from previous work by \cite{lipman2024flow}. we train our model on a MacBook Pro equipped with an Apple M3 Max chip and 48GB memory. This unified setup ensures consistent evaluation protocols ; each model is trained and evaluated across 10 random seeds, and we report the mean and standard deviation of all metrics. Codes and details around implementation and experiments are enclosed in supplementary material. 
We evaluate our proposed model, \textbf{UFM-TPP}, on multiple real-world benchmarks for long horizon marked event forecasting. In these experiments, we fix the number of future events to predict at \( n - c = 20 \), and extend to 10 and 5 in the ablation study. Since event sequences vary in total length, the context length \( c \) is adjusted accordingly. This setup aligns with the objective of long horizon multi-type event forecasting, where models must generate entire future trajectories rather than only the next event.

Our model is implemented in standard PyTorch, adapted from the flow matching framework of~\citet{lipman2024flow}. Training is conducted on a MacBook Pro with an Apple M3 Max chip and 48GB of memory. This unified setup ensures consistent evaluation protocols: each model is trained and evaluated over 10 random seeds, and we report the mean and standard deviation of all metrics. Code, implementation details, and additional experimental results are provided in the supplementary material.

%For fair comparison, we initialize all neural TPP baselines under the same randomization and hyperparameter selection protocol.

\paragraph{Datasets.}
We conduct experiments on six real-world multivariate event datasets spanning e-commerce, mobility, social media, and education. These datasets are curated and made available through the CDiff repository\footnote{\url{https://github.com/networkslab/cdiff}}.

\begin{itemize}
    \item \textbf{Taobao}~\citep{zhu2018learning}: User click sequences on a large-scale online recommendation platform.
    \item \textbf{Taxi}~\citep{whong-14-taxi}: Spatial trajectories of taxi pick-up events in New York City, binned by neighborhood.
    \item \textbf{StackOverflow}~\citep{snapnets}: Activity logs from a Q\&A platform, with events representing user interactions like posting and voting.
    \item \textbf{Retweet}~\citep{zhou2013learning}: Social cascades capturing retweet behaviors and user identities over time.
    \item \textbf{MOOC}~\citep{kumar2019predicting}: Learner interaction traces from online courses, including videos, quizzes, and forums.
    \item \textbf{Amazon}~\citep{amazon-2018}: Product review sequences, where each event corresponds to a user reviewing a category of item.
\end{itemize}

\paragraph{Baselines.}
We compare UFM-TPP with seven strong baselines, covering both autoregressive and non-autoregressive paradigms: 1). \textbf{Neural Hawkes Process (NHP)}~\citep{mei2016neural}: A recurrent neural point process that extends the classical Hawkes process to model both self-excitation and inhibition effects in event sequences.
    2). \textbf{Attentive NHP (AttNHP)}~\citep{yang2022transformer}: A multi-head attention variant of NHP that captures long-range dependencies.
    3). \textbf{LNM}~\citep{shchur2019intensity}: A log-normal mixture model that avoids explicit intensity modeling, allowing fast, intensity-free generation.
    4). \textbf{TCDDM}~\citep{linexploring}: A conditional denoising diffusion model for temporal point processes.
    5). \textbf{Dual-TPP}~\citep{deshpande2021long}: A hybrid temporal point process model that combines microscopic event dynamics and macroscopic count predictions through joint optimization to achieve accurate long horizon forecasting.
    % \item \textbf{HYPRO}~\citep{xue2022hypro}: A hybrid autoregressive method with candidate selection for multi-step forecasting.
    6). \textbf{HYPRO}~\citep{xue2022hypro}: A hybrid model that combines an autoregressive base predictor with a global energy function to reweight complete future sequences for improved long horizon forecasting.
    7). \textbf{CDiff}~\citep{zeng2024interacting}: The current state-of-the-art diffusion model for interacting timestamp and mark generation.

% \subsection{Evaluation Metrics}

% Following prior work~\citep{zeng2024interacting}, we assess long-horizon forecasting performance using both time-based and type-based metrics:

% \begin{itemize}
%     \item \textbf{OTD (Optimal Transport Distance)}~\citep{mei2019imputing}: Measures the cost of aligning the predicted event sequence with the ground truth using partial optimal matching.
%     \item \textbf{RMSE\textsubscript{x}}: Root Mean Squared Error of inter-event time predictions.
%     \item \textbf{RMSE\textsubscript{y}}: Root Mean Squared Error of event type predictions.
%     \item \textbf{sMAPE}: Symmetric Mean Absolute Percentage Error on predicted inter-event times.
% \end{itemize}

\paragraph{Evaluation Metrics.} Following prior work~\citep{zeng2024interacting}, we evaluate long horizon forecasting using 4 metrics that assess both the timing and type accuracy of predicted events.
1) \textbf{Optimal Transport Distance (OTD)}~\citep{mei2019imputing} captures sequence-level similarity by computing the minimum cost of aligning predicted and ground truth event streams, accounting for both timing and type mismatches. This metric is well-suited for evaluating multivariate sequences over long horizons.
2) \textbf{RMSE\textsubscript{x}} measures the average squared error in predicted inter-event times, emphasizing temporal precision. 3) \textbf{sMAPE} provides a normalized alternative, offering robustness across datasets with different time scales.
4) \textbf{RMSE\textsubscript{y}} evaluates event type prediction accuracy by comparing the predicted and true mark sequences at aligned positions.
Together, these metrics offer a comprehensive view of a model’s ability to jointly forecast when events happen and what types they are. %We report mean and standard deviation over 10 random seeds for all metrics.

\captionsetup[subfigure]{labelformat=empty}
\begin{figure*}[htbp]
    \centering
    % Top row: Inter-event time (dt) comparisons
    \begin{subfigure}[b]{0.33\textwidth}
        \centering
        \includegraphics[width=\textwidth]{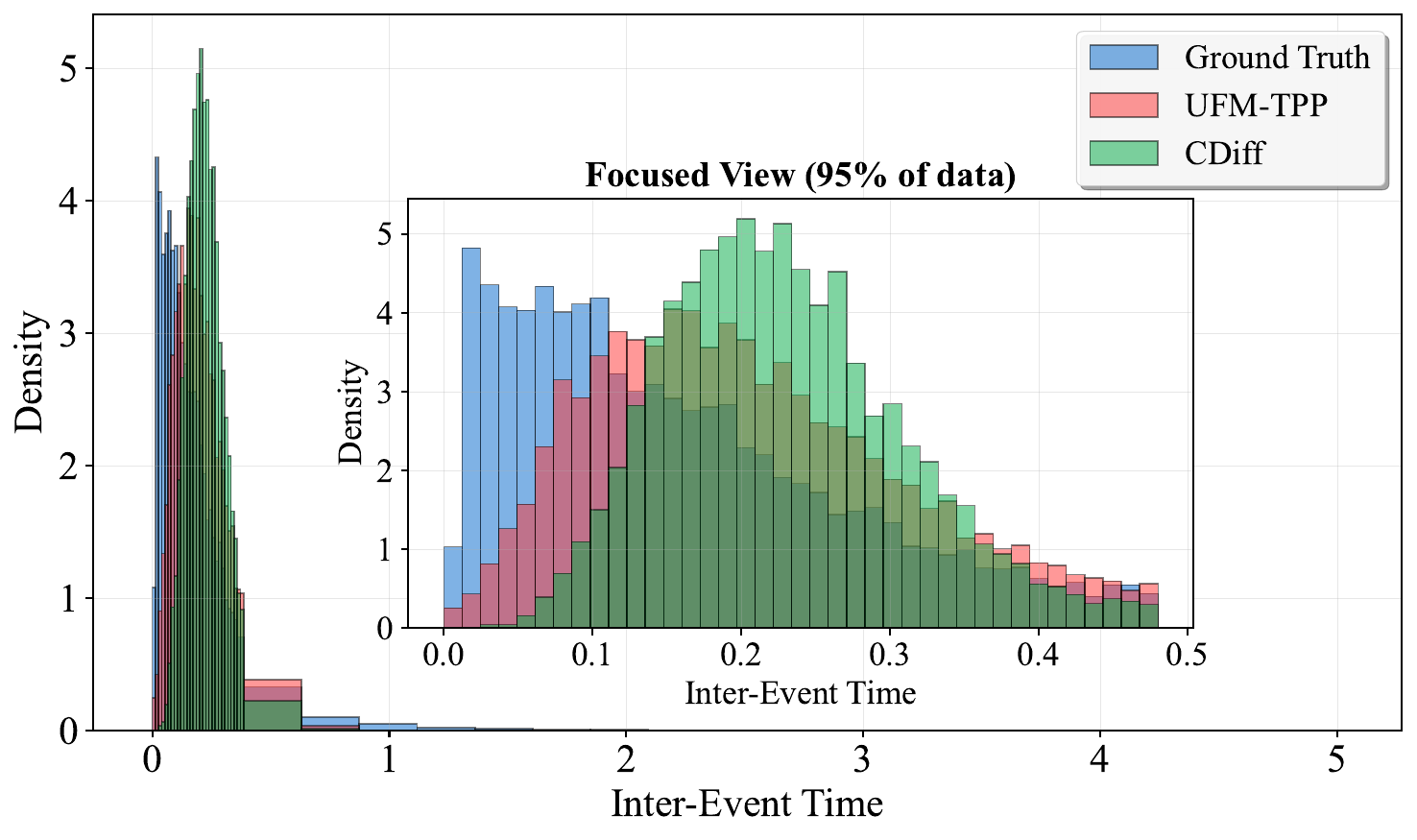}
        \caption{}
        \label{fig:taxi_dt}
    \end{subfigure}
    \hfill
    \begin{subfigure}[b]{0.33\textwidth}
        \centering
        \includegraphics[width=\textwidth]{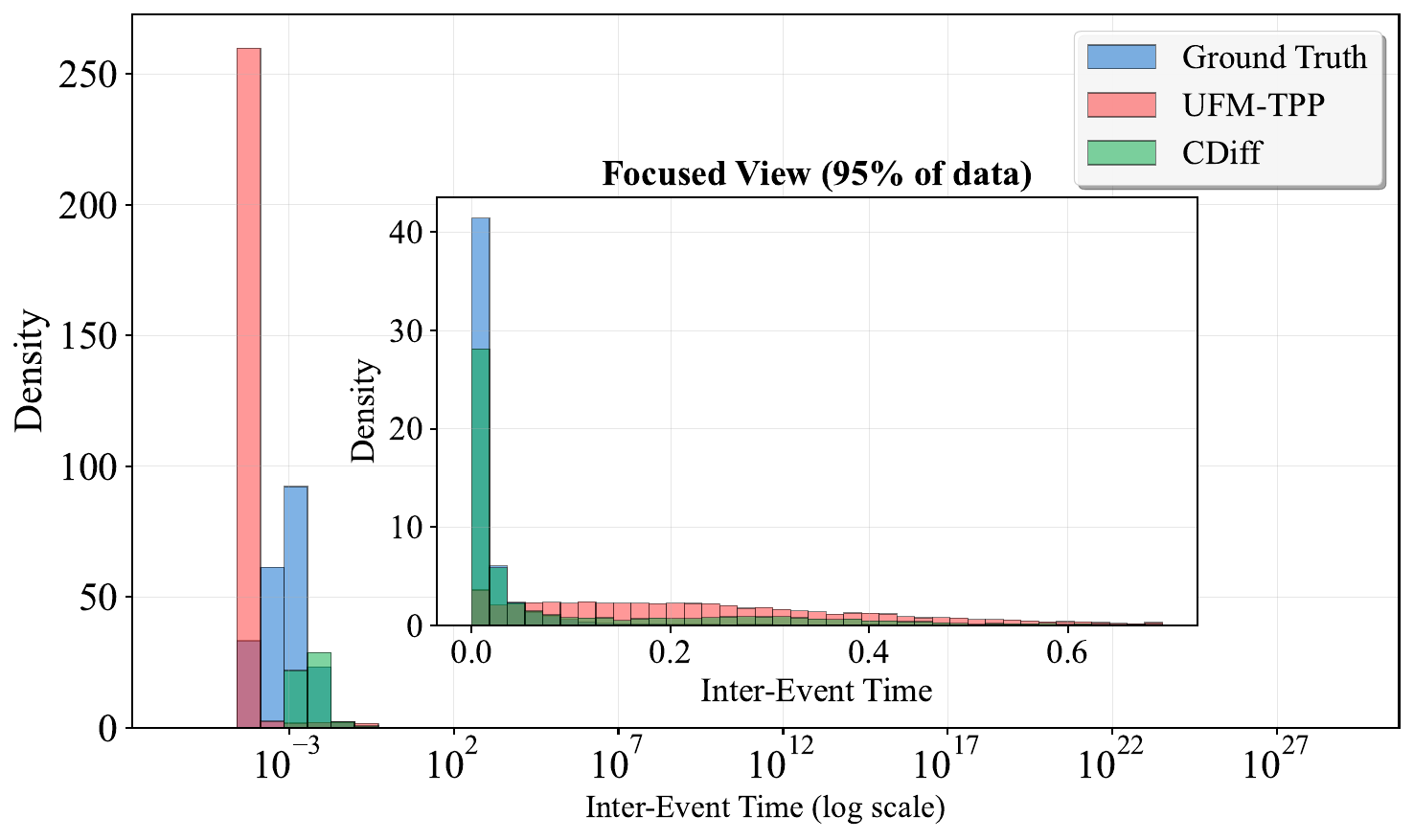}
        \caption{}
        \label{fig:taobao_dt}
    \end{subfigure}
    \hfill
    \begin{subfigure}[b]{0.33\textwidth}
        \centering
        \includegraphics[width=\textwidth]{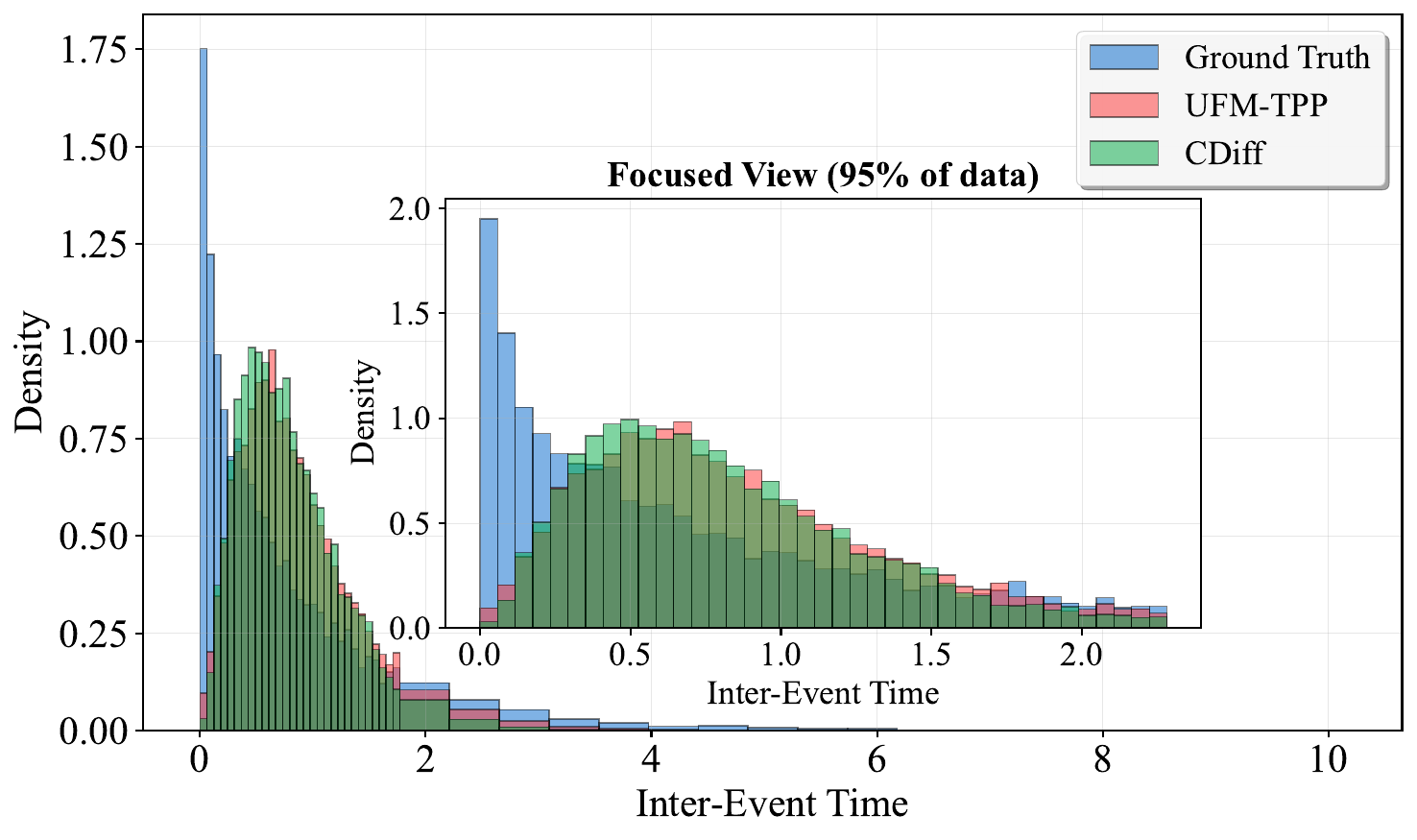}
        \caption{}
        \label{fig:sov2_dt}
    \end{subfigure}
    
    % \vspace{0.3cm}
    
    % Bottom row: Event type comparisons
    \begin{subfigure}[b]{0.33\textwidth}
        \centering
        \includegraphics[width=\textwidth]{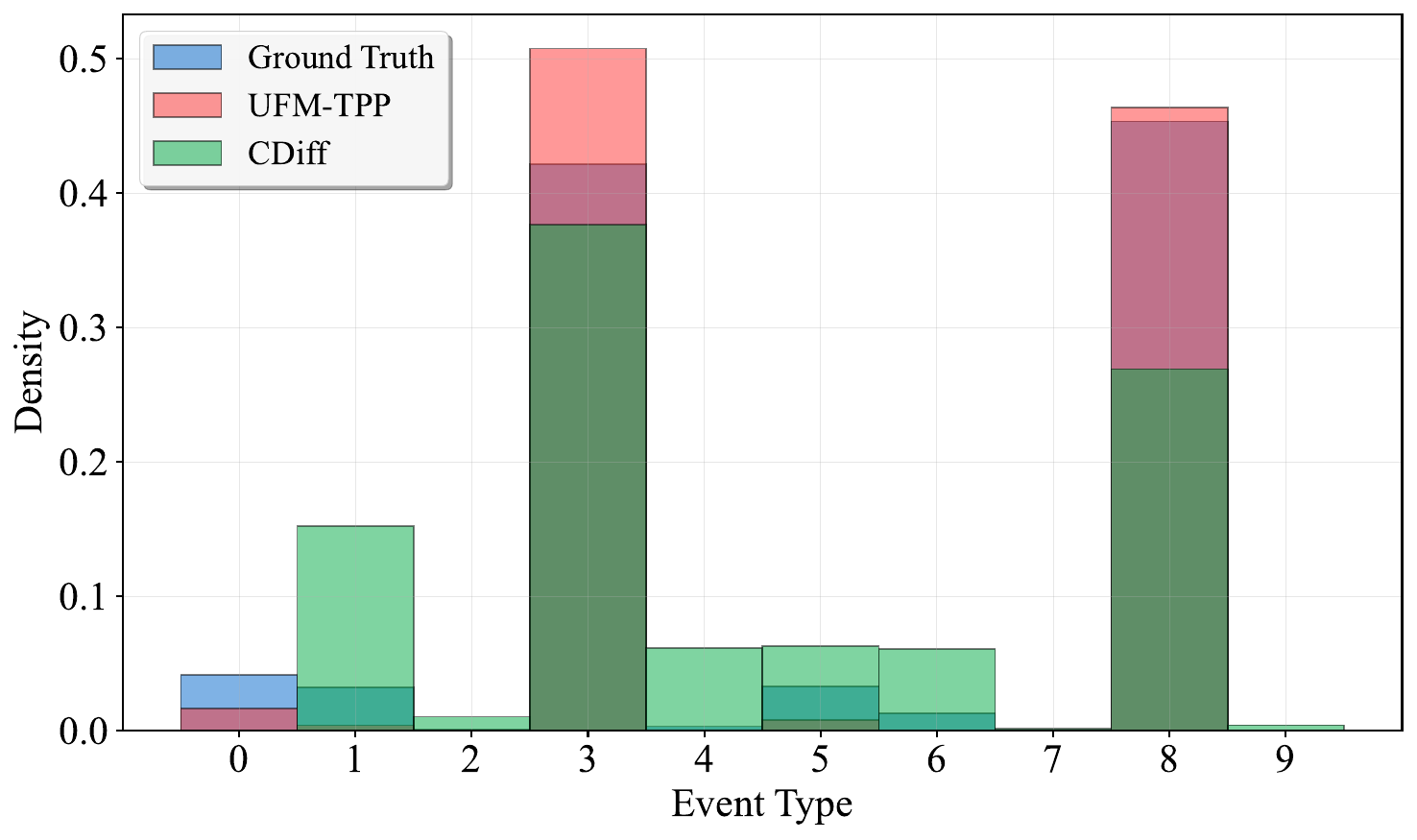}
        \caption{}
        \label{fig:taxi_type}
    \end{subfigure}
    \hfill
    \begin{subfigure}[b]{0.33\textwidth}
        \centering
        \includegraphics[width=\textwidth]{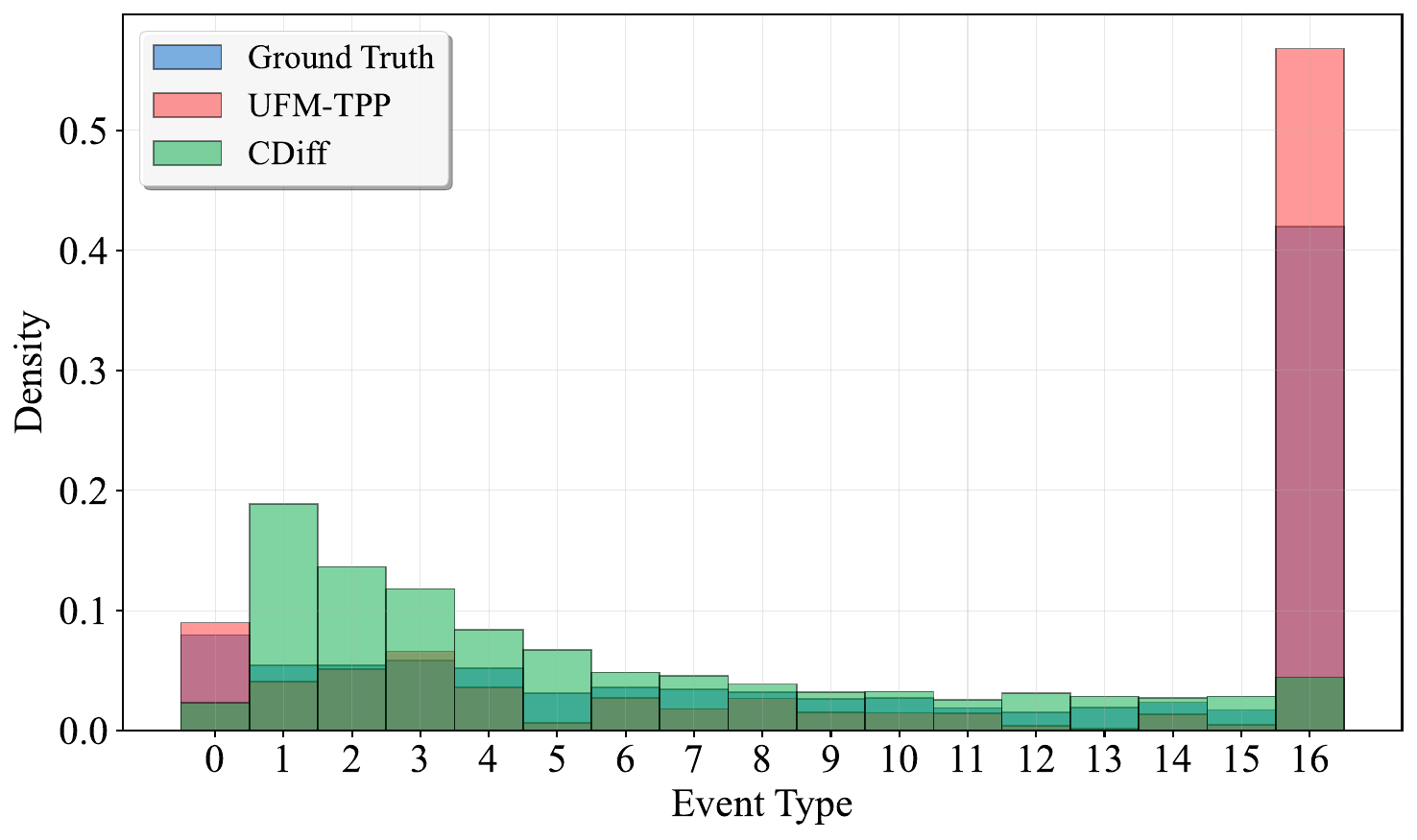}
        \caption{}
        \label{fig:taobao_type}
    \end{subfigure}
    \hfill
    \begin{subfigure}[b]{0.33\textwidth}
        \centering
        \includegraphics[width=\textwidth]{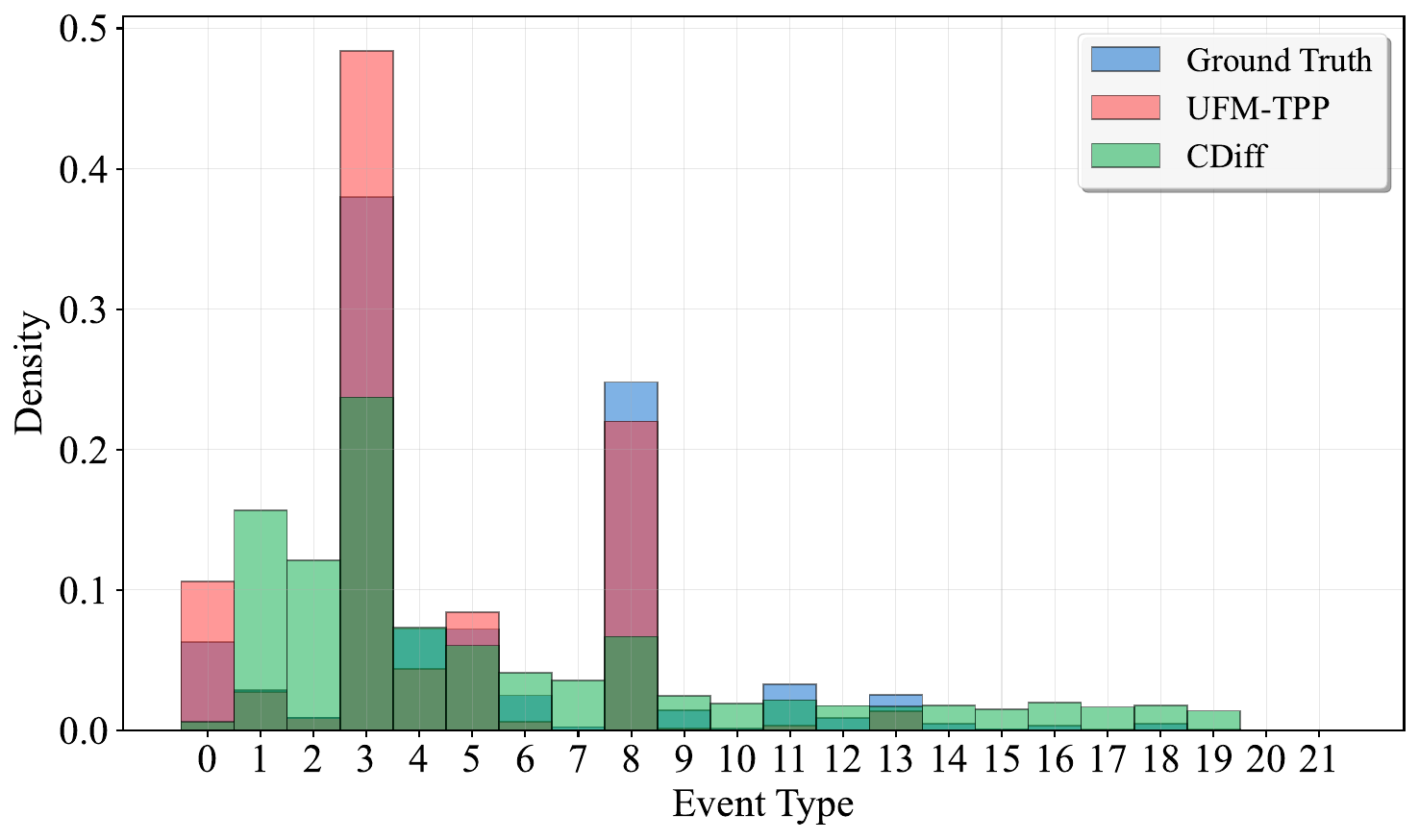}
        \caption{}
        \label{fig:sov2_type}
    \end{subfigure}
    \caption{Comparison of temporal point process models across three datasets. Top row shows inter-event time distributions and bottom row shows event type distributions. Each plot compares ground truth (blue), UFM-TPP predictions (red), and CDiff predictions (green) for Taxi, Taobao, and StackOverflow datasets (left to right). The insets in the top row provide focused views of the main distribution range, excluding extreme outliers for better visualization of central tendencies.}
    \label{fig:model_comparison}
\end{figure*}

\subsection{Main Results}
Table~\ref{tab:condensed_table} presents the forecasting performance of all methods across six real-world datasets. UFM-TPP demonstrates superior performance across most evaluation metrics, particularly excelling in sequence-level generation quality and interevent time prediction accuracy. Specifically, UFM-TPP achieves the best \textbf{OTD} on four out of six datasets, with notable improvements of 4.4\% on StackOverflow (39.439 vs 41.245) and 8.5\% on Taobao (40.552 vs 44.336), indicating superior ability to capture the overall distributional characteristics of event sequences. On the challenging Retweet dataset, TCDDM performs best, while CDiff achieves optimal performance on Taxi, demonstrating the competitive landscape among state-of-the-art methods. For time forecasting, UFM-TPP consistently delivers state-of-the-art $\textbf{RMSE}_{x}$ performance across five out of six datasets, with the exception of StackOverflow where CDiff achieves the best performance. UFM-TPP shows particularly strong temporal prediction capabilities on Taxi (0.298), Taobao (0.436), Retweet (22.647), MOOC (0.358), and Amazon (0.349), demonstrating the effectiveness of joint time-mark modeling through flow matching. The method achieves the best \textbf{sMAPE} scores on four datasets (Taxi, StackOverflow, Retweet, Amazon), with particularly impressive performance on StackOverflow showing a 16.2\% improvement (88.890 vs 106.175 compared to the second-best CDiff). On Amazon, UFM-TPP achieves the lowest sMAPE of 80.939, outperforming the second-best CDiff (81.987). %However, UFM-TPP shows weaker performance on Taobao sMAPE (161.212), where CDiff achieves the best result (125.685), and MOOC sMAPE (174.566), where HYPRO performs best (143.045). 
For mark prediction ($\textbf{RMSE}_{y}$), UFM-TPP shows more variable performance, achieving the best results on Taobao (2.169) and Amazon (1.931), while CDiff excels on Taxi (1.131) and MOOC (1.095), and Dual-TPP performs best on StackOverflow (1.134). This suggests that while the joint modeling approach excels at sequence-level coherence and interevent time prediction, specialized methods may have advantages for isolated mark prediction tasks. The consistent strong performance across diverse domains %from urban mobility (Taxi) to e-commerce (Taobao, Amazon), social networks (Retweet), educational platforms (MOOC), and developer communities (StackOverflow) 
demonstrates the robustness and generalizability of the flow matching framework for marked temporal point processes. These results validate our hypothesis that jointly modeling the evolution of both temporal and mark dimensions through unfied flow matching leads to more coherent and accurate event sequence generation compared to traditional autoregressive approaches and recent diffusion-based methods.%, particularly for applications where interevent time prediction accuracy and overall sequence fidelity are paramount.

\paragraph{Distribution Analysis.}
To provide deeper insights into the predictive capabilities of UFM-TPP, we analyze the distribution of generated inter-event times and event types on three representative datasets: Taxi, Taobao and StackOverflow, and compare its performance with CDiff as a close competitor. Figure~\ref{fig:model_comparison} presents comparative histograms showing the distributions of ground truth, UFM-TPP predictions, and CDiff predictions for both temporal and categorical dimensions. We include visual results of all datasets in the appendix. For inter-event time distributions, UFM-TPP demonstrates strong alignment with ground truth patterns across all datasets. On Taxi, both methods successfully capture the broad temporal distribution with good overlap in the main data range (0-0.5 inter-event durations), though UFM-TPP shows slightly better density estimation. The Taobao dataset reveals a critical advantage of UFM-TPP: while the ground truth exhibits a highly concentrated distribution near zero with long tails, UFM-TPP maintains numerical stability and produces reasonable predictions. In stark contrast, CDiff suffers from severe numerical instabilities on this dataset, generating extreme outlier values spanning 27 orders of magnitude as indicated by the log scale of inter-event time. On StackOverflow, both methods capture the characteristic short inter-event time distribution well.

% \begin{table*}[t!]
% \centering
% \caption{Performance metrics across different datasets and target lengths}
% \label{tab:length}
% \resizebox{0.85\textwidth}{!}{%
% \begin{tabular}{|l|l|cccccc|}
% \hline
% \textbf{Target Length} & \textbf{Metric} & \textbf{Taxi} & \textbf{Taobao} & \textbf{Stackoverflow} & \textbf{Retweet} & \textbf{Mooc} & \textbf{Amazon} \\
% \hline
% \multirow{4}{*}{10} & \textbf{OTD} & 11.311±0.191 & 19.284±0.305 & 20.430±0.209 & 31.096±0.340 & 22.208±0.464 & 18.358±0.231 \\
% % \cline{2-8}
% & $\textbf{RMSE}_y$ & 0.844±0.024 & 1.169±0.018 & 0.823±0.035 & 1.779±0.169 & 0.698±0.013 & 1.139±0.044 \\
% % \cline{2-8}
% & $\textbf{RMSE}_x$ & 0.318±0.010 & 0.430±0.015 & 1.052±0.033 & 24.223±0.669 & 0.317±0.020 & 0.330±0.009 \\
% % \cline{2-8}
% & \textbf{sMAPE} & 79.785±1.074 & 161.126±1.173 & 88.950±1.332 & 84.472±2.135 & 174.996±5.979 & 75.083±3.236 \\
% \hline
% \multirow{4}{*}{5} & \textbf{OTD} & 6.570±0.144 & 9.178±0.199 & 10.722±0.135 & 15.873±0.100 & 11.280±0.217 & 9.756±0.097 \\
% % \cline{2-8}
% & $\textbf{RMSE}_y$ & 0.623±0.023 & 0.629±0.010 & 0.556±0.019 & 1.159±0.073 & 0.439±0.006 & 0.701±0.014 \\
% % \cline{2-8}
% & $\textbf{RMSE}_x$ & 0.278±0.013 & 0.415±0.018 & 1.038±0.026 & 24.412±0.447 & 0.298±0.031 & 0.325±0.009 \\
% % \cline{2-8}
% & \textbf{sMAPE} & 78.669±2.146 & 153.331±4.761 & 87.354±1.436 & 81.368±1.411 & 178.724±4.022 & 76.370±2.389 \\
% \hline
% \end{tabular}%
% }
% \end{table*}

% 

UFM-TPP more accurately reproduces skewed event type distributions, particularly in datasets with dominant modes. On Taxi, UFM-TPP closely matches ground truth for types 3 and 8 (50.8\% vs. 42.2\% and 46.4\% vs. 45.4\%), while CDiff underestimates both and overallocates to rare types (e.g., type 1: 15.2\% vs. 3.2\%). On Taobao, UFM-TPP captures the peak at type 16 (56.8\% vs. 42.0\%), whereas CDiff flattens the distribution, misallocating mass to types 1 and 2. On StackOverflow, both models highlight dominant types, but UFM-TPP exhibits sharper mode concentration, while CDiff overspreads probability across infrequent types. These results highlight UFM-TPP’s strength in modeling peaked distributions, while CDiff tends to over-diversify at the cost of fidelity. These distributional analyses underscore UFM-TPP’s key strength in numerical stability and temporal accuracy. %while also revealing limitations in capturing event type diversity. 
Notably, UFM-TPP delivers consistent and accurate predictions across all datasets, in contrast to CDiff, which suffers from severe numerical failures on challenging cases like Taobao which highlights the critical importance of algorithmic robustness for practical deployment.
\begin{figure}[tbhp]
    \centering
    \includegraphics[width=0.49\textwidth]{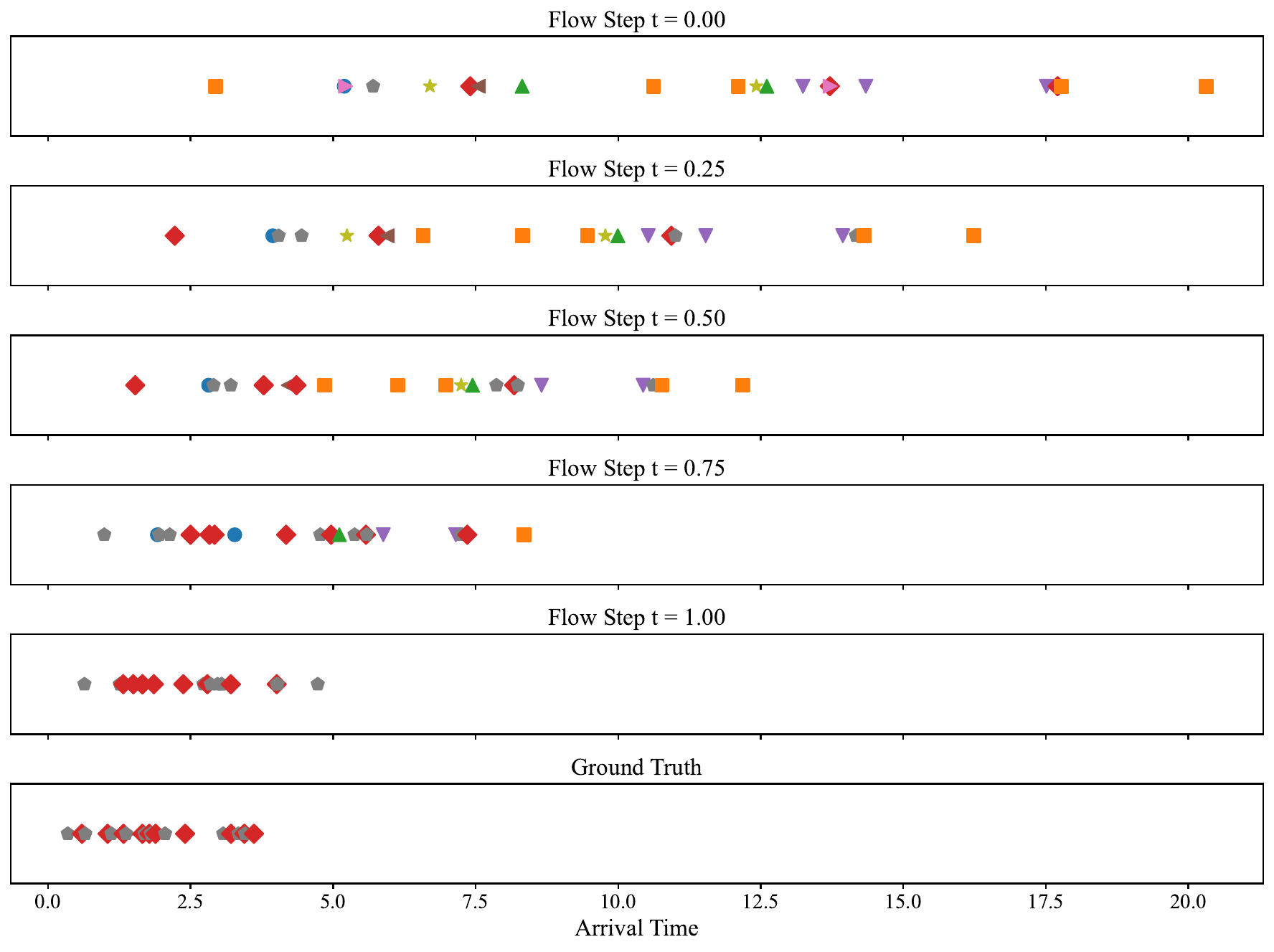}
    \caption{%
   Flow Matching Generation Process for Temporal Point Processes.
        Evolution of event sequences through key flow steps showing the transformation from random noise to realistic event patterns. 
        Each subplot displays events positioned by their arrival times, with different colors and markers representing event types from taxi dataset.  
        The flow progresses from $t=0.00$ (initial random state) through intermediate steps at $t=0.25$, $t=0.50$, and $t=0.75$, 
        culminating at $t=1.00$ (final generated sequence). 
        The bottom panel shows the ground truth sequence for comparison. 
        % The model successfully learns to generate coherent temporal patterns that closely match the target distribution, 
        % demonstrating the effectiveness of joint time-mark modeling through continuous normalizing flows.
        %Range values indicate the temporal span of each sequence.
    }
    \label{fig:flow_generation_evolution}
\end{figure}

% \begin{figure}[htbp]
%     \centering
%     \includegraphics[width=0.48\textwidth]{figures/taxi_arrival_tmes.pdf}
%     \caption{Evolution of event sequences through flow steps. The figure shows the progressive generation of event sequences at different time steps (t = 0.00, 0.25, 0.50, 0.75, 1.00) compared to the ground truth. Each marker represents an event with different shapes and colors indicating different event types.}
%     \label{fig:flow_steps_evolution}
% \end{figure}

\subsection{Complexity via Run Time}
We compare the computational efficiency of our proposed model, UFM-TPP, against the diffusion-based baseline CDiff across six benchmark datasets. Both models are trained for 100 epochs under their optimal hyperparameter settings using Apple’s M3 chip with MPS acceleration. As shown in Table~\ref{tab:performance}, UFM-TPP achieves significantly faster sampling times across all datasets. For instance, on the Taxi dataset, UFM-TPP requires only 7 seconds to generate samples, compared to 86 seconds for CDiff—a 12× speedup. Similar improvements are observed across other datasets, such as Taobao (16s vs. 204s) and Amazon (12s vs. 175s). This efficiency stems from UFM-TPP’s design, which requires fewer than 10 flow steps for generation while maintaining competitive accuracy (an example is shown in Fig. \ref{fig:flow_generation_evolution}). In contrast, CDiff relies on a few hundred iterative diffusion steps, resulting in much higher inference latency. Although UFM-TPP has a larger parameter count (150K–155K vs. 42K–58K for CDiff), its training times are consistently lower across all datasets. On Taobao, for example, UFM-TPP completes training in 343 seconds, nearly 4× faster than CDiff's 1322 seconds. This performance advantage highlights the practical scalability of UFM-TPP, making it a compelling choice for both training and deployment in long horizon event forecasting scenarios.

\begin{table}[h]
\centering
\caption{Performance Comparison Across Different Datasets}
\label{tab:performance}
\resizebox{\columnwidth}{!}{%
\footnotesize
\setlength{\tabcolsep}{3pt}
\begin{tabular}{l|l|cccccc}
\hline
\textbf{Model} & \textbf{Metric} & \textbf{Taxi} & \textbf{Taobao} & \textbf{Stack} & \textbf{Retweet} & \textbf{Mooc} & \textbf{Amazon} \\
\hline
\multirow{3}{*}{CDiff} & Sampling (s) & 86 & 204 & 173 & 179 & 402 & 175 \\
& Params (K) & 42.8 & 55.8 & 56.2 & 54.6 & 58.5 & 55.7 \\
& Training (s) & 955 & 1322 & 1272 & 6388 & 4677 & 5042 \\
\hline
\multirow{3}{*}{UFM-TPP} & Sampling (s) & 7 & 16 & 20 & 23 & 137 & 12 \\
& Params (K) & 150.2 & 151.1 & 151.8 & 149.3 & 155.4 & 151.0 \\
& Training (s) & 122 & 343 & 882 & 5020 & 3867 & 3386 \\
\hline
\end{tabular}
}
\end{table}

\begin{table}[htbp]
\centering
\caption{Performance metrics across different datasets. Upper panel: target length 10; Lower panel: target length 5}
\label{tab:length}
\large % Increased font size
\renewcommand{\tabcolsep}{1.5pt} % Decreased column spacing (default is 6pt)
\resizebox{\columnwidth}{!}{%
\begin{tabular}{l|cccccc}
\hline
\textbf{Metric} & \textbf{Taxi} & \textbf{Taobao} & \textbf{Stackoverflow} & \textbf{Retweet} & \textbf{Mooc} & \textbf{Amazon} \\
\hline
\textbf{OTD} & 11.311±0.191 & 19.284±0.305 & 20.430±0.209 & 31.096±0.340 & 22.208±0.464 & 18.358±0.231 \\
% \hline
$\textbf{RMSE}_y$ & 0.844±0.024 & 1.169±0.018 & 0.823±0.035 & 1.779±0.169 & 0.698±0.013 & 1.139±0.044 \\
% \hline
$\textbf{RMSE}_x$ & 0.318±0.010 & 0.430±0.015 & 1.052±0.033 & 24.223±0.669 & 0.317±0.020 & 0.330±0.009 \\
% \hline
\textbf{sMAPE} & 79.785±1.074 & 161.126±1.173 & 88.950±1.332 & 84.472±2.135 & 174.996±5.979 & 75.083±3.236 \\
\hline
\end{tabular}%
}

\vspace{0.5em}

\resizebox{\columnwidth}{!}{%
\begin{tabular}{l|cccccc}
\hline
\textbf{Metric} & \textbf{Taxi} & \textbf{Taobao} & \textbf{Stackoverflow} & \textbf{Retweet} & \textbf{Mooc} & \textbf{Amazon} \\
\hline
\textbf{OTD} & 6.570±0.144 & 9.178±0.199 & 10.722±0.135 & 15.873±0.100 & 11.280±0.217 & 9.756±0.097 \\
% \hline
$\textbf{RMSE}_y$ & 0.623±0.023 & 0.629±0.010 & 0.556±0.019 & 1.159±0.073 & 0.439±0.006 & 0.701±0.014 \\
% \hline
$\textbf{RMSE}_x$ & 0.278±0.013 & 0.415±0.018 & 1.038±0.026 & 24.412±0.447 & 0.298±0.031 & 0.325±0.009 \\
% \hline
\textbf{sMAPE} & 78.669±2.146 & 153.331±4.761 & 87.354±1.436 & 81.368±1.411 & 178.724±4.022 & 76.370±2.389 \\
\hline
\end{tabular}%
}
\end{table}

\subsection{Ablation Study}
To further evaluate the effectiveness of UFM-TPP across different forecasting horizons, we conduct an ablation study with additional target lengths (i.e., 5 \& 10) as shown in 
Table \ref{tab:length}. Comparing UFM-TPP's performance across different target lengths reveals a clear trend of improved accuracy with shorter prediction horizons, with substantial improvements observed across all evaluation metrics and all six datasets. This performance scaling aligns with the general expectation that shorter-term predictions are inherently more accurate due to reduced uncertainty propagation, confirming that UFM-TPP's architecture effectively scales across different forecasting horizons while maintaining its efficiency.

% Alternative: Landscape orientation for wider tables
% \begin{sidewaystable}
% \centering
% \caption{Performance metrics across different datasets and target lengths}
% \label{tab:performance_metrics}
% \begin{tabular}{|l|l|c|c|c|c|c|c|}
% [same content as above but will be rotated 90 degrees]
% \end{tabular}
% \end{sidewaystable}

\section{Conclusion}
% In this work, we introduced Unified Flow Matching for Marked Temporal Point Processes (UFM-TPP), a novel framework for efficient and unified long-horizon multi typed event forecasting. Our method jointly models the evolution of inter-event times and event types through coupled continuous and discrete flow dynamics, bypassing the limitations of traditional autoregressive models and recent diffusion-based approaches. By leveraging flow matching principles and conditioning on encoded context histories, UFM-TPP enables non-autoregressive generation of future event sequences with strong fidelity to both temporal and categorical structure. While our method shows strong performance, it assumes independence among future events given the context, which may limit its expressiveness for highly structured sequences conditioned on the history context. Future work will explore relaxing this assumption and incorporating uncertainty quantification for more flexible and robust event generation.
In this work, we introduced Unified Flow Matching for Marked Temporal Point Processes (UFM-TPP), a novel framework for efficient and unified long horizon forecasting of marked event sequences. Our approach jointly models the evolution of inter-event times and event types using coupled continuous and discrete flow dynamics, overcoming limitations of traditional autoregressive models and recent diffusion-based methods. By leveraging flow matching and conditioning on encoded context histories, UFM-TPP enables non-autoregressive generation of future event sequences with high fidelity to both temporal and categorical structure. While the model demonstrates strong performance, it assumes conditional independence among future events given the context, which may limit expressiveness for highly structured sequences. Future work will focus on relaxing this assumption to support more flexible and robust event generation.

\end{document}